\documentclass{article}

\PassOptionsToPackage{numbers, compress}{natbib}

\usepackage[preprint]{neurips_2025}

\usepackage[utf8]{inputenc} 
\usepackage[T1]{fontenc}    
\usepackage{hyperref}       
\usepackage{url}            
\usepackage{booktabs}       

\usepackage{amsfonts}       

\usepackage{graphicx}       
\usepackage{nicefrac}       
\usepackage{microtype}      
\usepackage{xcolor}     
\usepackage{amsmath}
\usepackage{siunitx}
\usepackage{multirow} 
\usepackage{geometry}
\usepackage[linesnumbered,ruled,vlined]{algorithm2e}
\usepackage{wrapfig}

\usepackage[linesnumbered,ruled,vlined]{algorithm2e}
\usepackage{xcolor}
\usepackage{tcolorbox}
\usepackage{amsmath}
\usepackage{tcolorbox}
\usepackage{cleveref}       
\tcbuselibrary{breakable}

\definecolor{myblue}{RGB}{20, 75, 160}
\definecolor{mygreen}{RGB}{0, 140, 70}
\definecolor{myred}{RGB}{200, 30, 45}
\definecolor{myorange}{RGB}{255, 128, 0}

\SetKwComment{Comment}{$\triangleright$\ }{}
\SetKwInput{KwIn}{\textcolor{myblue}{Input}}
\SetKwInput{KwOut}{\textcolor{myblue}{Output}}

\SetAlgoNlRelativeSize{-1}
\SetAlFnt{\small}
\SetKwProg{Fn}{Function}{:}{}
\SetKwFunction{FMain}{SelfGuidedTTS}

\title{Guided by Gut: Efficient Test-Time Scaling\\with Reinforced Intrinsic Confidence}

\author{%
  Amirhosein Ghasemabadi \\
  ECE Department, University of Alberta\\
  \texttt{ghasemab@ualberta.ca} \\
  \And
  Keith G. Mills \\
  ECE Department, University of Alberta\\
  \texttt{kgmills@ualberta.ca} \\
  \And
  Baochun Li \\
  ECE Department, University of Toronto\\
  \texttt{bli@eecg.toronto.edu} \\
  \And
  Di Niu \\ 
  ECE Department, University of Alberta\\
  \texttt{dniu@ualberta.ca}
}

\begin{document}

\maketitle

\begin{abstract}
Test-Time Scaling (TTS) methods for enhancing Large Language Model (LLM) reasoning often incur substantial computational costs, primarily due to extensive reliance on external Process Reward Models (PRMs) or sampling methods like Best-of-N (BoN). This paper introduces Guided by Gut (GG), an efficient self-guided TTS framework that achieves PRM-level performance without costly external verifier models. Our method employs a lightweight tree search guided solely by intrinsic LLM signals—token-level confidence and step novelty. One critical innovation is improving the reliability of internal confidence estimates via a targeted reinforcement learning fine-tuning phase. Empirical evaluations on challenging mathematical reasoning benchmarks demonstrate that GG enables smaller models (e.g., 1.5B parameters) to achieve accuracy matching or surpassing significantly larger models (e.g., 32B–70B parameters), while reducing GPU memory usage by up to 10×. Compared to PRM-based methods, GG achieves comparable accuracy with 8× faster inference speeds and 4–5× lower memory usage. Additionally, GG reduces KV cache memory usage by approximately 50\% compared to the BoN strategy, facilitating more efficient and practical deployment of TTS techniques. 

\smallskip
\noindent\small The code is available at \url{https://github.com/Amirhosein-gh98/Guided-by-Gut}.
\end{abstract}

\section{Introduction}

\begin{wrapfigure}{rt}{0.5\textwidth}
    \centering
    \includegraphics[width=2.8in]{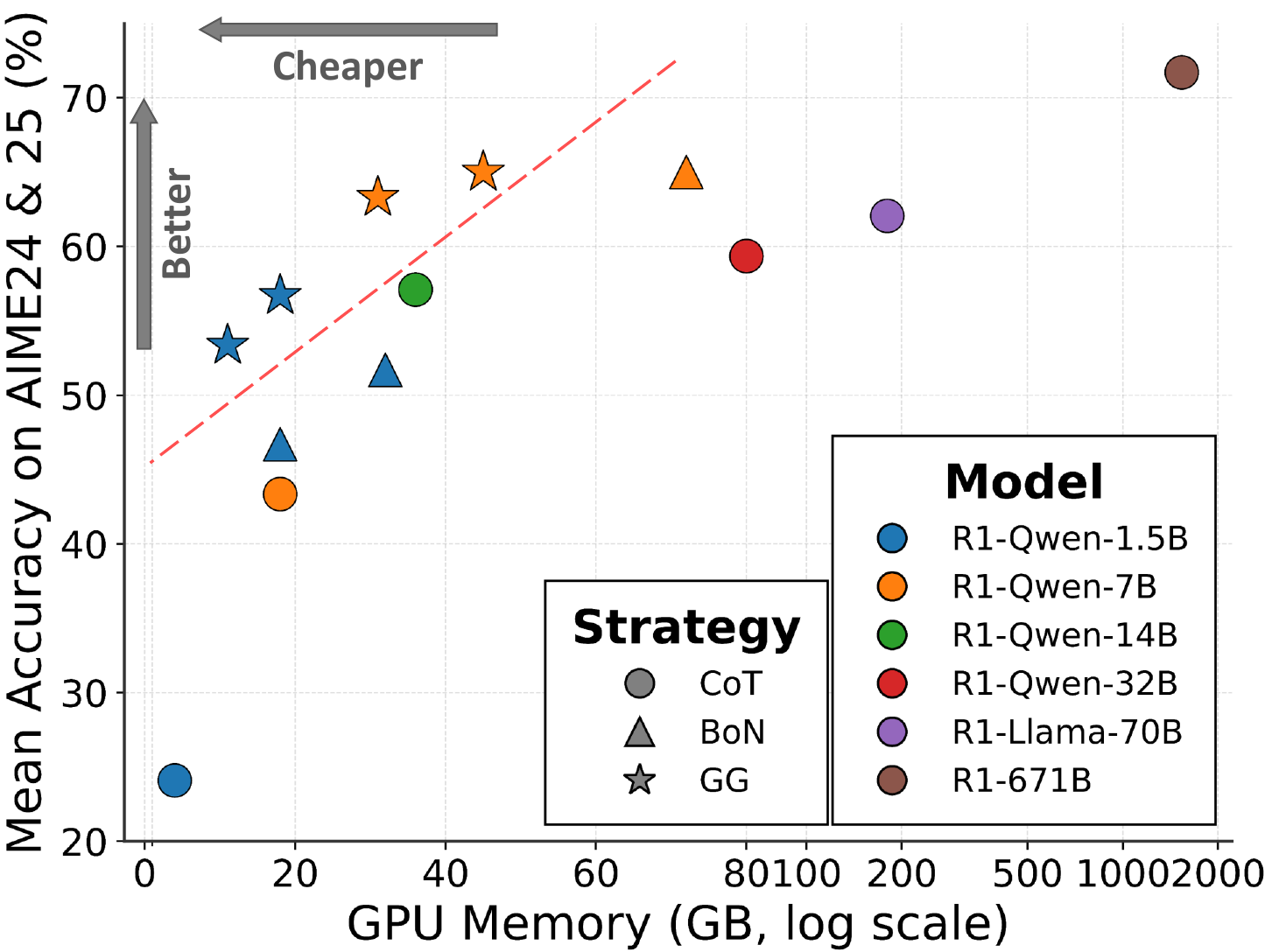} 
    \caption{We compare the performance and GPU VRAM usage of Guided by Gut (GG; stars) to Best-of-N (BoN; triangles) and Chain-of-Thought (CoT; circles) on several LLMs. GG achieves better accuracy at much lower memory cost (log-scaled).} 
    \label{fig:intro_pareto}
    \vspace{-28mm}  
\end{wrapfigure}

Enhancing the performance of Large Language Models (LLMs) often requires significant computational resources through model scaling~\cite{achiam2023gpt,ouyang2022training, villalobos2022will} or complex inference strategies~\cite{ji2025test,zhou2024survey}. Test-Time Scaling (TTS) techniques like Chain-of-Thought (CoT)~\cite{wei2022chain} 
allocate 
additional computation during inference. This re-allocation of compute resources provides 
a powerful alternative for boosting LLM reasoning capabilities, as evidenced by models like OpenAI's o series~\cite{openai2024reasoning}, DeepSeek R1~\cite{guo2025deepseek}, and others 
~\cite{qwen2.5,team2025kimi, bai2025qwen2,muennighoff2025s1}.

Contemporary TTS methods~\cite{lightman2023let, wang2023math, snell2024scaling} are capable of enhancing LLMs containing 1.5B parameters 
such that they outperform 70B, 405B parameter or even large closed-source LLMs on difficult reasoning and mathematical benchmarks~\cite{liu2025can}. However, TTS is an expensive search process where the total compute cost to generate an answer matches or may even exceed that of a larger LLM~\cite{zhang2025lessons, luo2025deepscaler}. 
For example, Sampling-based methods~\cite{wang2023math} like Best-of-N (BoN)~\cite{brown2024large} operate by generating a large number of candidate solutions (e.g., potentially hundreds) and then choosing the optimal one from this pool, which requires prohibitively large amounts of LLM inference for complex tasks. In addition, Process Reward Models 
are auxiliary verification models which guide the TTS process by providing step-by-step correctness feedback~\cite{xiong2024implementation,zheng2024processbench, zhang2025lessons, wang2023math}. Such verifier-guided techniques can be computationally expensive to train and deploy and suffer from generalizability issues~\cite{liu2025can,zhong2025comprehensive, zheng2024processbench}.
Thus, regardless of strategy, TTS for small-scale LLMs relies on expensive inference,  
which severely limits practical application and motivates the need for more cost-effective TTS frameworks. 

To bridge this gap, we propose Guided by Gut (GG), a computationally efficient and scalable TTS framework to enhance LLM reasoning. GG leverages intrinsic signals derived from the LLM's generation process, fine-tuned by reinforcement learning (RL), to enable smaller models to achieve substantially stronger reasoning performance which matches or exceeds results achieved by much larger models (e.g., 70B) and expensive TTS strategies at much lower GPU memory costs, as Figure~\ref{fig:intro_pareto} illustrates. Our detailed contributions are as follows.

\begin{itemize}
\setlength{\itemsep}{0pt}
\item \textbf{Token Confidence and Novelty:} Instead of relying on costly external verifier models, we utilize intrinsic cues from the LLM output, e.g., token probabilities, which we interpret as confidence scores and measure the novelty of a potential reasoning step. This provides a lightweight avenue for guiding inference-time search and can be integrated into existing models and algorithms. 

\item \textbf{Reinforced Confidence via RL Fine-tuning:} We incorporate RL via Group Relative Policy Optimization (GRPO)  
into model fine-tuning specifically to improve the reliability of LLM internal confidence estimation, leading to more reliable guidance for our test-time search strategy.
\item \textbf{Efficient Test-Time Search with Self-Guidance:} We introduce a tree search algorithm based on Diverse Verifier Tree Search (DVTS)~\cite{beeching2024scalingtesttimecompute} guided by the LLM's intrinsic signals (token probability/confidence, novelty). To achieve efficient TTS, our algorithm is specifically optimized for minimal computational cost during inference. 
\end{itemize}

We apply GG to reasoning LLMs from DeepSeek R1 family~\cite{guo2025deepseek} and Qwen2.5-Math~\cite{yang2024qwen2} as a non-reasoning model and evaluate it on benchmark tasks like AIME24/25~\cite{AIME24}, MATH500~\cite{MATH}, and AMC~\cite{AMC23}. Experimental results not only demonstrate that GG achieves significant performance improvements over relevant baselines such as BoN and CoT, but also highlight its superior computational efficiency. Specifically, GG enables smaller models (e.g., 1.5B-7B parameters) to outperform much larger counterparts (e.g., 32B and 70B), achieving similar or superior accuracy while using up to 4×–10× less GPU memory. Furthermore, compared to computationally expensive PRM-based approaches, GG achieves comparable accuracy at a fraction of the computational cost, leading to 4×–5× lower GPU memory usage and up to 8× faster inference speeds. Furthermore, GG achieves an approximately 50\% reduction in KV cache memory usage compared to the BoN strategy, facilitating significantly more efficient and cost-effective deployment of reasoning LLMs.

\begin{figure}[t!]
    \centering
    \includegraphics[width=1\textwidth]{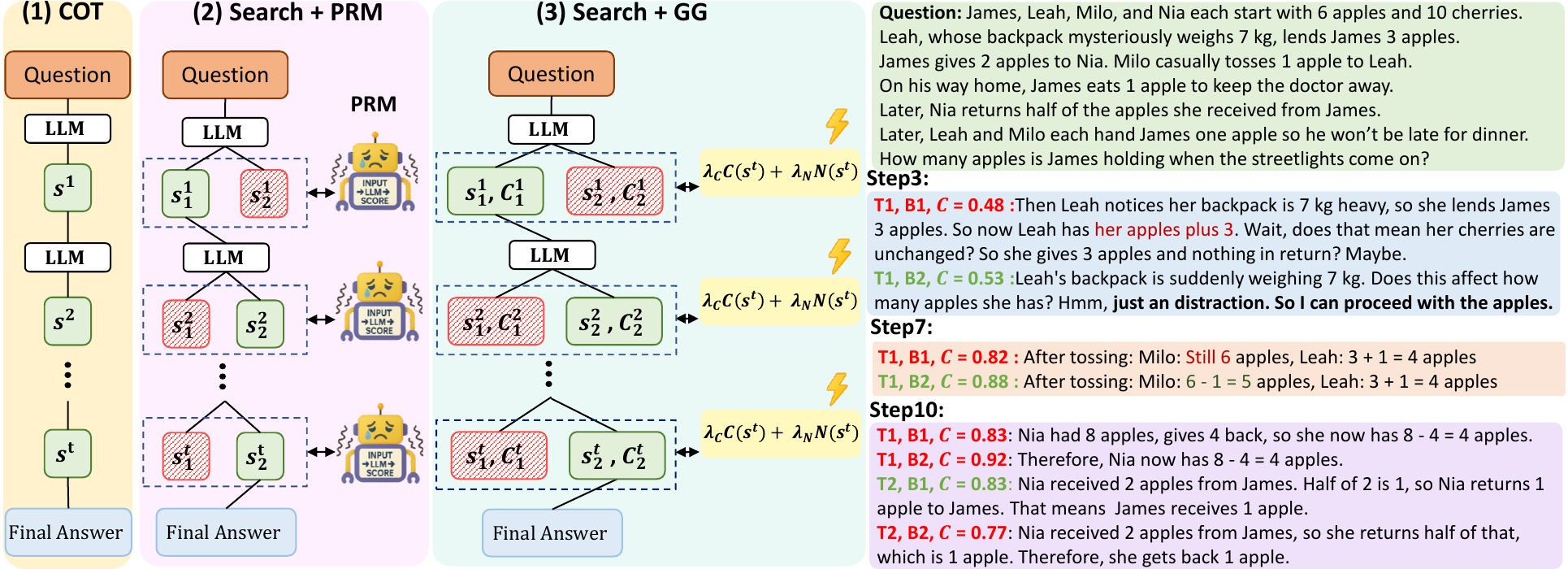} 
    \caption{Comparison of reasoning generation strategies. 
             \textbf{(1)} Standard Chain-of-Thought (CoT) generates a single reasoning path autoregressively. 
             \textbf{(2)} Search guided by an external Process Reward Model (PRM) explores multiple candidate steps ($s_1^t, s_2^t, \dots$), using PRM scores to select promising paths. 
             \textbf{(3)} Our proposed Self-Guided Search similarly explores multiple steps but uses intrinsic signals, Confidence ($\mathcal{C}$) and Novelty ($N$), derived from the LLM to guide the search at each step without relying on an external PRM. In the example, \(\mathbf{T}\) stands for an independent \textbf{tree} and \(\mathbf{B}\) stands for a \textbf{branch} within that tree. Example text best read zoomed-in.}
    \label{fig:reasoning_strategies}
\end{figure}

\section{Related Work}
\label{sec:related}

\textbf{Test-Time Scaling (TTS).}
OpenAI's o1~\cite{openai2024reasoning} highlighted the critical role of enhanced reasoning for complex tasks like code generation and mathematical problem solving. 
The subsequent open release of models like DeepSeek-R1~\cite{guo2025deepseek} further spurred research into reasoning mechanisms like Chain-of-Thought (CoT)~\cite{wei2022chain}. CoT falls under the umbrella of Test-Time Scaling ~\cite{beeching2024scalingtesttimecompute, face2025open}, which enhances model performance by strategically allocating additional computational effort during inference, through methods like extended reasoning or search, rather than relying solely on increased model size or longer training.

As we illustrate in Figure~\ref{fig:reasoning_strategies}, CoT primarily consists of letting the LLM generate reasoning steps and reach an answer/conclusion in a purely autoregressive manner. 
Sampling methods like Best-of-N (BoN)~\cite{brown2024large} involve generating multiple independent solutions, from which the optimal one is selected, often with the aid of a verifier or through voting mechanisms. Tree-based methods like Beam Search~\cite{xie2023self}, Diverse Verifier Tree Search (DVTS)~\cite{beeching2024scalingtesttimecompute}, and Monte Carlo Tree Search (MCTS)~\cite{xie2024monte} provide different algorithmic avenues towards selecting optimal multi-step reasoning paths. While these models enable $<$10B parameter reasoning LLMs to perform at a level similar to much larger models, e.g., 70B and above~\cite{liu2025can}, they necessitate multiple rounds of inference to generate large quantities of potential reasoning steps or chains. 

\textbf{External Verification.} To guide exploration towards the most promising step or reasoning path, many Test-Time Scaling methods require a mechanism to quantify the effectiveness of different choices. This role is typically filled by an external verifier~\cite{snell2024scaling, beeching2024scalingtesttimecompute}, such as a Process Reward Model (PRM)~\cite{liu2025can} or an Outcome Reward Model (ORM)~\cite{lightman2023let}. These verifiers are often substantial models themselves, significantly contributing to the overall computational burden of TTS. While various strategies like confidence-weighted voting~\cite{taubenfeld2025confidence, razghandi2025cer}, dynamic early stopping~\cite{wan2024dynamic}, and sample pruning~\cite{huang2023fewer} aim to mitigate this overhead, many sophisticated search-based TTS approaches still fundamentally rely on a powerful (and thus costly) PRM or similar verifier to effectively guide the search. For instance, even approaches combining model confidence with PRM supervision still ultimately rely on the costly PRM~\cite{xie2023self}. To address this, GG avoids costly external verifiers for simple, near-zero overhead internal signals, proving this minimalist approach surprisingly effective.

\textbf{Reinforcement Learning for LLM Reasoning.}
Recent literature highlights Reinforcement Learning's crucial role in advancing Large Language Model reasoning without human intervention~\cite{face2025open}. ReFT\cite{luong2024reft} employs Proximal Policy Optimization (PPO) to enhance the generalizability of LLMs for reasoning. A key algorithm, Group Relative Policy Optimization (GRPO)~\cite{shao2024deepseekmath}, notably eliminates the need for a separate value function in PPO. Further research explores various RL training aspects to improve reasoning capabilities~\cite{yu2025dapo, zeng2025simplerl,liu2025understanding}. DeepScaleR~\cite{luo2025deepscaler} aims to boost existing reasoning models' through additional GRPO fine-tuning with iterative context lengthening.

\newcommand{\Conf}{\operatorname{Conf}}
\section{Methodology}
\label{sec:method}
This section outlines our proposed method, Guided by Gut (GG). We begin by providing essential background on the Test-Time Scaling process. Following this, we elaborate on the self-guided search mechanism and overall strategy. 

\subsection{Preliminaries}
\paragraph{Problem Formulation.}
Given an input prompt or question $Q$, our objective is to generate a logical reasoning chain $R = [s^1, s^2, \dots, s^T]$ leading to a correct final answer $A$, where each step $s^t$ typically constitutes a sentence or short paragraph incrementally building upon previous steps. The overall reasoning process thus follows the pipeline:
\begin{align*}
Q \rightarrow R \rightarrow A,
\end{align*}
with the reasoning chain $R$ explicitly bridging the input question and the final answer through intermediate logical steps.

\paragraph{Chain-of-Thought (CoT) Reasoning.}
Standard Chain-of-Thought (CoT)~\cite{wei2022chain} approaches jointly generate the reasoning chain and final answer via autoregressive language modeling. Formally, given $Q$, the model sequentially generates each reasoning step conditioned on previously generated steps:
\begin{align}
\label{eq:cot}
P(R = s^{1:T} \mid Q) = \prod_{t=1}^{T} P(s^t \mid Q, s^{1:t-1}).
\end{align}
Each step $s^t$ thus depends on the input $Q$ and preceding reasoning steps $s^{1:t-1}$, mirroring standard autoregressive token generation in language modeling.

\paragraph{Guiding Search with Reward Models.}
Single-path autoregressive generation methods can suffer from error accumulation~\cite{wu2025more, mukherjee2025premise}. To mitigate this, tree search methods explore multiple reasoning trajectories simultaneously. These methods typically use an external Process Reward Model (PRM) or Outcome-supervised Reward Model (ORM) for step-wise correctness evaluations. A PRM is a model that, given an input $Q$ and previous steps $s^{1:t-1}$, assigns a correctness score or reward $r_t$ to candidate next steps $s^t$:

\begin{align}
\label{eq:reward_combined}
r_t = \operatorname{PRM}(s^{t} \mid Q, s^{1:t-1}) \quad \text{or} \quad r_t = \operatorname{PRM}(s^{t} \mid Q).
\end{align}

Likewise, an ORM is a sparse reward model where only the final step receives a non-zero reward; $r_{t<T}=0$. Thus, these reward models improve logical coherence and accuracy by guiding search algorithms like Beam Search, BoN and DVTS. 

\subsection{Proposed Method: Self-Guided Search}

The usage of verifier models like PRMs and ORMs, while effective, introduces computational overhead and generalizability issues~\cite{liu2025can}. 
To address these limitations, we propose \textbf{Guided by Gut (GG)}, which leverages the 
intrinsic signals directly obtained from the LLM internal token 
generation process. This removes the dependency on external evaluation, ensuring minimal computational overhead. Specifically, our approach uses two intrinsic signals to guide reasoning:
\begin{itemize}

\item \textbf{Confidence} $C(s^t)$ reflects the internal assurance a model has with respect to a 
given reasoning step $s^t$. We compute confidence 
directly 
from token-level probabilities:
\begin{equation}
    \centering
    C(s^t) = \frac{1}{m_t}\sum_{l=1}^{m_t} \log p(s^t_{l} \mid \text{context}),
    \label{eq:confidence}
\end{equation}
where $m_t$ is the number of tokens in reasoning step $s^t$ and `context' represents previous tokens in $s^t$ and all prior reasoning steps. 
\item \textbf{Novelty} $N(s^t)$ encourages exploration by measuring the dissimilarity of candidate reasoning steps to previously explored paths. Specifically, we calculate novelty as the proportion of new tokens introduced by the candidate step $s^t$ relative to tokens already explored within the current reasoning context. 
\end{itemize}

We formulate a reward $r_t$ to guide the search process by combining these intrinsic signals as follows: 

\begin{equation}
    \centering
    r_t = \lambda_C C(s^t) + \lambda_N N(s^t),
    \label{eq:gg_reward}
\end{equation}
where $\lambda_N$ and $\lambda_C$ balance exploration and exploitation, respectively. Unlike verifier-guided approaches, our reward is intrinsically computed from LLM prediction statistics, eliminating external dependency.

\subsection{Enhancing Confidence via Reinforcement Learning Fine-Tuning}
\label{sec:method_grpo}

A significant challenge in using intrinsic statistics 
is ensuring 
reliability, as raw model confidence and answer novelty may not accurately reflect correctness. To refine this process, 
we incorporate a Reinforcement Learning fine-tuning phase.

Specifically, we 
utilize Group Relative Policy Optimization (GRPO)~\cite{shao2024deepseekmath}, a memory-efficient variant of Proximal Policy Optimization (PPO)~\cite{schulman2017proximal} tailored for LLM applications. Let $\pi$ represent the 
LLM we want to fine-tune,  
parameterized either current fine-tuned weights $\theta$, fine-tuned weights from the previous iteration $\theta_{old}$ or the original reference weights $\theta_{ref}$. 
At each iteration 
GRPO samples a group of $G$ outputs $\{o_i\}_{i=1}^G$ from 
$\pi_{\theta_{\text{old}}}$, where each output $o_i$ represents a chain of reasoning steps and an answer $o_i = [R_i, A_i]$. Each output receives a reward $r_i$, which we describe below:

\begin{figure*}[t!]
    \centering
    \includegraphics[width=1\textwidth]{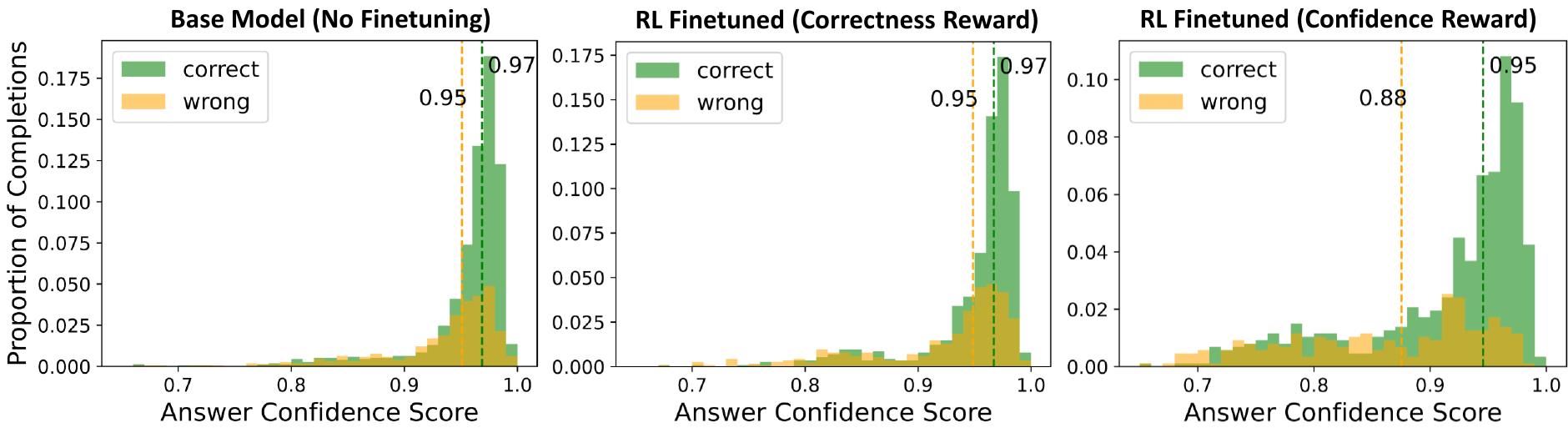}
    \caption{
        \textbf{Answer Confidence Distribution Across Training Settings.}
        Each subplot shows the normalized distribution of confdence scores for correct (green) and incorrect (orange) completions across different fine-tuning strategies.
        The vertical dashed lines mark the mean confidence for correct and wrong completions, respectively.
        The \textbf{base model} (left) is generally overconfident, with incorrect completions receiving high confidence scores.
        \textbf{Fine-tuning with correctness reward} (middle) improves accuracy but leaves the confidence distribution largely unchanged.
        \textbf{Confidence-based fine-tuning} (right) better separates correct from incorrect completions, showing improved calibration.
    }
        \label{fig:confidence-distribution}
\end{figure*}

\paragraph{Confidence-Based Reward.}
Our RL training enhances confidence accuracy with a novel reward function, moving beyond typical correctness-only rewards. Many literature methods use such rewards for the final answer, yielding sparse, binary signals which provide no learning signal when no completion is correct. In contrast, our fine-tuning reward is more comprehensive: it specifically integrates the model's internal confidence throughout reasoning steps with final answer correctness. This richer, multifaceted signal is crucial for calibrating model certainty, enabling more reliable self-guided search. 

To compute it, we first calculate a weighted summation of the confidence scores for the last $k$ reasoning steps 
in the reasoning chain $R_i$, $\mathcal{C}(R_i)$: 

\begin{equation}
    \centering
    \mathcal{C}(R_i) = \frac{1}{\sum_{l=1}^{k} l} \sum_{l=1}^{k} l \cdot c(s^{T-k+l}_i) 
    \label{eq:seq_confidence}
\end{equation}

Then the RL fine-tuning reward $r_i$ is computed based on $A_i$'s correctness and the reasoning chain confidence $\mathcal{C}(R_i)$, as follows:
\begin{equation}
    \centering
    r_i =
    \begin{cases}
    1 + \mathcal{C}(R_i)^4 & \text{if } \operatorname{IsCorrect}(A_i), \\
    1 - 10\mathcal{C}(R_i)^4 & \text{otherwise},
    \end{cases}
    \label{eq:conf_reward}
\end{equation}

where
$\operatorname{IsCorrect}(A_i)$ returns a boolean validating the final answer as correct or not. 
Equation~\ref{eq:conf_reward} ensures that correct, highly confident answers are rewarded more strongly, whereas incorrect, overconfident answers receive greater penalties, thus promoting precise confidence calibration. 

The calibration effect is demonstrated in Figure~\ref{fig:confidence-distribution}.

\paragraph{Advantage and Fine-tuning Update.}

After computing the fine-tuning reward $r_i$ for each sampled output $o_i$, we compute the normalized advantage $\hat{A}_i$ as follows: 

\begin{equation}
    \centering
    \hat{A}_{i} = \frac{r_i - \operatorname{mean}(\{r_j\}_{j=1}^G)}{\operatorname{std}(\{r_j\}_{j=1}^G)}.
    \label{eq:grpo_advantage}
\end{equation}

We can then compute the clipped surrogate policy update $\delta_i$~\cite{schulman2017proximal} for a given output as

\begin{equation}
    \centering
    \delta_i = \dfrac{1}{|o_i|}\sum_{l=1}^{|o_i|}[\text{min}(r_i \hat{A}_i, \text{clip}(r_i, 1-\epsilon, 1+\epsilon)\hat{A}_i)-\beta\mathbb{D}_{KL}(\pi_{\theta}|\pi_{ref})],
    \label{eq:clipped_surrogate}
\end{equation}

where $\epsilon$ controls the magnitude of the update and the Kullback-Leibler divergence term $\mathbb{D}_{KL}$, controlled by $\beta$, prevents the fine-tuning weights from moving too far away from the original weights. We then update the LLM by using all clipped surrogates to maximize the GRPO objective.

\begin{equation}
    \centering
    \mathcal{J}_{\text{GRPO}}(\theta) = \mathbb{E}_{
    \{o_i\}_{i=1}^G \sim \pi_{\theta_{\text{old}}}}[ \dfrac{1}{G}\sum_{i=1}^G \delta_i].
    \label{eq:grpo_update}
\end{equation}

\subsection{Search Strategy}

We employ Diverse Verifier Tree Search (DVTS)~\cite{beeching2024scalingtesttimecompute}, an extension of beam search that splits initial beams into independent subtrees that are expanded greedily using a reward model or our proposed intrinsic rewards. DVTS 
enhances solution diversity and performance. 
We apply DVTS within the CoT framework by operating at the reasoning step ($s^t$) level, identifying step completions via model-specific delimiters. This structure allows the search to evaluate and expand complete logical increments during the reasoning process. 

Our modified DVTS algorithm runs recursively until it encounters one of the termination conditions we specify: a maximum reasoning depth or a maximum token length limit is exceeded, the reasoning exhibits signs of text degeneration, e.g., excessive repetition, or in the best case, the LLM arrives at a final answer $A$. To avoid incomplete outputs due to overthinking near the limits, we inject a model-specific signal prompting a conclusion. For example, with DeepSeek models, appending ``**Final Answer**'' effectively elicits the final answer, ensuring usable completions even in complex cases. For further details on DVTS, we provide a formal explanation in the supplementary materials due to space constraints.

\section{Experimental Results}
\label{sec:exp}
In this section, we validate our Guided by Gut Test-Time Scaling (GG) approach. 
We first describe our experimental setup, 
then present our findings. Finally, we ablate the components of our method. 

\subsection{Confidence Calibration via Reinforcement Learning}
We perform a reinforcement learning fine-tuning phase.
Our primary goal during RL fine-tuning is not to maximize raw task accuracy but to enhance the reliability of the intrinsic confidence signals utilized by GG through GRPO, as described in Section~\ref{sec:method_grpo} and in our reward function (Equation~\ref{eq:conf_reward}).

For this calibration step, we utilize the LIMO dataset~\cite{ye2025limoreasoning}. This dataset contains 817 high-quality examples curated specifically for complex mathematical reasoning, featuring detailed, structured solutions. We fine-tune the \texttt{DeepSeek-R1-Distill-Qwen-1.5B} \& \texttt{7B} models with Low-Rank Adapters (LoRA)~\cite{hu2022lora}. 
The implementation leverages the \texttt{TRL}~\cite{vonwerra2022trl} library and an adaptation of the \texttt{open-r1} codebase~\cite{face2025open}. Key aspects of the setup include a low learning rate of $2.0 \times 10^{-6}$ with a cosine scheduler~\cite{loschchilov2017cosine}, LoRA rank $r=128$ and $\alpha=128$, and GRPO training with $G=8$ generations per prompt at a temperature of $0.6$. We perform fine-tuning on two NVIDIA A100 80 GPUs using 
\texttt{bfloat16} precision with FlashAttention-2~\cite{dao2023flashattention} optimization, using prompt/completion length limits of 768/8096 tokens and a batch size of 8 per device with 8 gradient accumulation steps for 3 epochs. The fine-tuning process completes in approximately one day.

\subsection{Results}
\begin{table*}[!t]
\centering
\caption{Performance metrics for various models and TTS strategies on the AIME24 and AIME25 benchmarks. Inference Speed and GPU Memory measured using NVIDIA A100 80GB cards.}
\label{tab:model_performance_metrics_custom_order}
\setlength\tabcolsep{3.5pt} 
\resizebox{\textwidth}{!}{%
\begin{tabular}{@{}llcccc@{}} 
\toprule
\textbf{Model} & 
\textbf{TTS Strategy} & 
\begin{tabular}[c]{@{}c@{}}\textbf{AIME24}\\\scriptsize Acc. [\%]$\uparrow$\end{tabular} &
\begin{tabular}[c]{@{}c@{}}\textbf{AIME25}\\\scriptsize Acc. [\%]$\uparrow$\end{tabular} &
\begin{tabular}[c]{@{}c@{}}\textbf{Inference Speed}\\\scriptsize time/question [m] $\downarrow$\end{tabular} &
\begin{tabular}[c]{@{}c@{}}\textbf{GPU Memory}\\\scriptsize [GB]$\downarrow$\end{tabular} \\
\midrule
DeepSeek-R1-Distill-Qwen-1.5B & CoT
    & 26.8
    & 21.4 
    & 0.2 
    & 4 \\
DeepSeek-R1-Distill-Qwen-7B & CoT
    & 48.1 
    & 38.6 
    & 1.0
    & 18 \\
DeepSeek-R1-Distill-Qwen-14B & CoT
    & 65.8 
    & 48.4 
    & 6.5 
    & 36 \\
DeepSeek-R1-Distill-Qwen-32B & CoT
    & 66.9 
    & 51.8 
    & 11.5 
    & 80 \\
DeepSeek-R1-Distill-Llama-70B & CoT
    & 70.0   
    & 54.1 
    & 20.0   
    & 180  \\
DeepSeek-R1-671B & CoT
    & 79.1 
    & 64.3 
    & --   
    & 1536 \\
OpenAI o1 mini & CoT
    & 63.6 
    & --   
    & --   
    & -- \\
\midrule
DeepSeek-R1-Distill-Qwen-1.5B & BoN($N=32$)
    & 56.7 
    & 36.7 
    & 2.8  
    & 18 \\
DeepSeek-R1-Distill-Qwen-1.5B & BoN($N=64$)
    & 66.7 
    & 36.7 
    & 5.1  
    & 32 \\
\midrule 
DeepSeek-R1-Distill-Qwen-1.5B & GG ($N=32$)
    & 66.7 
    & 40.0   
    & 2.7  
    & 11 \\ 
DeepSeek-R1-Distill-Qwen-1.5B & GG ($N=64$)
    & 66.7 
    & 46.7 
    & 5.1  
    & 18 \\
DeepSeek-R1-Distill-Qwen-7B & GG ($N=32$)
    & 73.3 
    & 53.3 
    & 10.3 
    & 31 \\
DeepSeek-R1-Distill-Qwen-7B & GG ($N=64$)
    & 76.7 
    & 53.3 
    & 18.0   
    & 45 \\
\bottomrule
\end{tabular}%
} 
\end{table*}

To demonstrate the efficacy and efficiency of Guided by Gut (GG), we first benchmark our approach using <10B parameter LLMs against several larger Chain-of-Thought (CoT) LLMs, some of which are closed-source like OpenAI o1 mini. We also compare to competitive tree-based TTS methods such as Best-of-N (BoN), a robust TTS baseline that outperforms PRM-guided search with R1 models~\cite{liu2025can}, using standardized evaluation settings.

As for the evaluation settings, we allocate token budgets based on method requirements: 16k tokens for tree-based TTS methods, i.e., GG and BoN, and 32k tokens for standard CoT models (consistent with original papers) to balance comprehensive comparisons with computational feasibility. We set the maximum reasoning steps to 200 for R1 models. We configure BoN to use $N=32$ or $N=64$ samples using majority voting. Likewise, we evaluate our Confidence-Guided DVTS with equivalent compute budgets: using $N=32$ total paths (from 16 subtrees with $M=2$ beam width/verifiers) and $N=64$ total paths (from 32 subtrees with $M=2$), distinctively employing weighted majority voting based on final answer confidence scores.

We evaluate GG 
on the AIME 2024 \& 2025 (AIME24, AIME25)~\cite{AIME24, aime25} benchmarks, which each consist 
of 30 problems 
from the respective American Invitational Mathematics Examinations that 
emphasize advanced high-school-level reasoning. 
Table~\ref{tab:model_performance_metrics_custom_order} presents our findings. 

First, we observe that GG delivers competitive performance compared to much larger LLMs employing CoT strategies. Specifically, a 1.5B parameter LLM with GG achieves accuracy comparable to a 32B CoT model, while a 7B GG LLM reaches accuracy levels closer to those of a 70B CoT model—and even beats OpenAI’s o1 mini on AIME24. Moreover, the results on \texttt{DeepSeek-R1-Distill-Qwen-7B} are equally impressive. Even with a moderate sampling budget ($N=32$), GG outperforms all CoT-based LLMs with fewer than 100B parameters. Notably, it achieves this while requiring at most \textit{one sixth} of the VRAM memory and delivering faster inference compared to \texttt{Distill-Llama-70B}, despite producing similar accuracy. Only the \texttt{DeepSeek-R1-671B} model, which has nearly 10× the parameters, achieves higher accuracy on both benchmarks—but at a prohibitively high memory cost, requiring nearly \textit{30×} more VRAM memory.

On the \texttt{DeepSeek-R1-Distill-Qwen-1.5B} model, GG consistently outperforms BoN (at $N=32$ and $N=64$) while using almost 50\% less GPU memory. 
For instance, with a computational budget of $N=32$, GG attains 10\% and 3.3\% performance on AIME24 and AIME25, respectively. Further, for $N=64$, GG outperforms BoN by 10\% on AIME25 (46.7\% vs. 36.7\%). In summary, Table~\ref{tab:model_performance_metrics_custom_order} demonstrates that GG is a competitive, self-guided TTS method that allows smaller models to achieve accuracy comparable to larger counterparts while using significantly less memory.

\paragraph{Comparison with Process Reward Models.} 
We further compare the self-directed search mechanism of GG to TTS guided by an external verifier, i.e., a Process Reward Model (PRM). We conduct this experiment using \texttt{Qwen2.5-Math-1.5B-Instruct}~\cite{bai2025qwen2} as the base model. This is a non-reasoning LLM commonly employed for evaluating PRM-guided TTS in the literature~\cite{wang2024openr, beeching2024scalingtesttimecompute, liu2025can}. Specifically, we set a limit of 4k tokens and 50 reasoning steps per trial for this experiment, which reflects the shorter non-CoT answers and the overhead of PRM-guided search. 

We evaluate on two datasets: AMC2023 (AMC23)~\cite{zwhe99_amc23_dataset}, comprising 40 problems from the American Mathematics Competition testing foundational high-school math skills; and MATH500, a diverse set of 500 problems randomly sampled from the full MATH benchmark~\cite{lightman2023let}.

Table~\ref{tab:prm_vs_no_prm} summarizes our findings from comparing GG, which utilizes simple intrinsic search signals, against external PRM-guided approaches. Our aim is to evaluate whether a lightweight, cost-effective signal could achieve competitive reasoning performance. On AMC23, GG consistently outperforms both the math-shepherd-mistral-7B-PRM (by 5.0\%) and the RLHFlow/Llama3.1-8B-PRM (by 3\%) across both $N$ settings. On MATH500, GG's performance is also highly competitive, trailing the PRM accuracies by a narrow margin of only 0.1 to 1.1 percentage points. Crucially, GG achieves this strong standing with its significantly more lightweight methodology: it is ~8x faster and uses <5GB of VRAM, as it avoids loading an additional large reward model. These results underscore GG's robust performance and efficiency, especially for local deployment scenarios.

\begin{table*}[!t]
\centering
\caption{Performance comparison of PRM-based and non-PRM (GG) scoring strategies using \texttt{Qwen2.5-Math-1.5B-Instruct} as the base LLM with DVTS as the search algorithm. Horizontal line demarcates $N=16$ from $N=32$ total path hyperparameter. We denote the external verifier utilized in PRM trials. We vary the total paths $N$ for both methods and the external verifier utilized in PRM trials. 
Maximum token limit of 4048. Higher accuracy and lower inference speed/memory are preferred. Best and second best results in \textbf{bold} and \textit{italics}, respectively.} 
\label{tab:prm_vs_no_prm}
\setlength\tabcolsep{4pt}
\resizebox{\textwidth}{!}{%
\begin{tabular}{@{}lcccc@{}}
\toprule
\textbf{Scoring Strategy} &
\begin{tabular}[c]{@{}c@{}}\textbf{AMC23}\\\scriptsize Acc. [\%]$\uparrow$\end{tabular} &
\begin{tabular}[c]{@{}c@{}}\textbf{MATH500}\\\scriptsize Acc. [\%]$\uparrow$\end{tabular} &
\begin{tabular}[c]{@{}c@{}}\textbf{Inference Speed}\\\scriptsize time/question [m] $\downarrow$\end{tabular} &
\begin{tabular}[c]{@{}c@{}}\textbf{GPU Memory}\\\scriptsize [GB]$\downarrow$\end{tabular} \\

\midrule

PRM (math-shepherd-mistral-7B-PRM, $N=16$) 
    & 58.3 
    & \textbf{79.1} 
    & \textit{0.8} 
    & \textit{19}\\

PRM (RLHFlow/Llama3.1-8B-PRM, $N=16$) 
    & \textit{60.0}
    & \textit{79.0}
    & 0.9 
    & 21\\

Guided by Gut (GG; ours) ($N=16$) 
    & \textbf{63.3} 
    & 78.9
    & \textbf{0.1} 
    &  \textbf{4}\\
\midrule
PRM (math-shepherd-mistral-7B-PRM, $N=32$) 
    & 60.0 
    & \textbf{81.1} 
    & \textit{1.5} 
    & \textit{23}\\

PRM (RLHFlow/Llama3.1-8B-PRM, $N=32$) 
    & \textit{61.7} 
    & \textit{80.7} 
    & 1.6 
    & 25 \\

Guided by Gut (GG; ours) ($N=32$) 
    & \textbf{65.0} 
    & 80.0 
    & \textbf{0.2} 
    &  \textbf{5}\\
    
\bottomrule
\end{tabular}%
}
\end{table*}

\begin{table*}[!t]
\centering
\caption{Accuracy and KV Cache memory usage for various models and TTS strategies across mathematical reasoning datasets. Horizontal lines demarcate base model and search budget. Accuracy reported as mean (max) for AIME24/25 and mean performance for MATH500/AMC. Higher is better for accuracy; lower is better for KV Cache memory.}
\label{tab:tts_accuracy_final}
\setlength\tabcolsep{4pt} 
\resizebox{\textwidth}{!}{
\begin{tabular}{@{}llccccc@{}}
\toprule
\textbf{Model} & \textbf{TTS Strategy} &
\begin{tabular}[c]{@{}c@{}}\textbf{AIME24}\\\scriptsize Acc. [\%]$\uparrow$\end{tabular} &
\begin{tabular}[c]{@{}c@{}}\textbf{AIME25}\\\scriptsize Acc. [\%]$\uparrow$\end{tabular} &
\begin{tabular}[c]{@{}c@{}}\textbf{MATH500}\\\scriptsize Acc. [\%]$\uparrow$ \end{tabular} &
\begin{tabular}[c]{@{}c@{}}\textbf{AMC}\\\scriptsize Acc. [\%]$\uparrow$ \end{tabular} &
\begin{tabular}[c]{@{}c@{}}\textbf{KV Cache}\\\scriptsize [GB]$\downarrow$\end{tabular} \\
\midrule

\multirow{6}{*}{DeepSeek-R1 1.5B~\cite{guo2025deepseek}}
  & CoT
    & 26.8
    & 21.4
    & 83.9
    & 68.3
    & 0.86 \\
\cmidrule(lr){2-7}

  & BoN ($N=32$)
    & 52.2(56.7)
    & 34.4(36.7)
    & 91.7
    & 90.0
    & 13.7 \\

  & GG ($N=32$)
    & 58.9(66.7)
    & 37.5(40.0)
    & 91.9
    & 92.5
    & 6.9 \\
\cmidrule(lr){2-7}

  & BoN ($N=64$)
    & 57.8(66.7)
    & 36.7(36.7)
    & 92.3
    & 90.0
    & 27.4 \\

  & GG ($N=64$)
    & 61.7(66.7)
    & 42.2(46.7)
    & 92.9
    & 93.3
    & 13.7 \\
\midrule

\multirow{6}{*}{DeepSeek-R1 7B~\cite{guo2025deepseek}}
  & CoT
    & 48.1
    & 38.6
    & 92.8
    & 85.5
    & 1.7 \\
\cmidrule(lr){2-7}

  & BoN ($N=32$)
    & 72.2(73.3)
    & 51.1(53.3)
    & 96.1
    & 92.5
    & 27.2 \\

  & GG ($N=32$)
    & 71.7(73.3)
    & 52.5(53.3)
    & 96.3
    & 94.1
    & 13.7 \\
\cmidrule(lr){2-7}

  & BoN ($N=64$)
    & 75.5(76.7)
    & 50.0(53.3)
    & 96.1
    & 93.7
    & 54.4 \\
    
  & GG ($N=64$)
    & 76.7(76.7)
    & 51.7(53.3)
    & 96.5
    & 95.0
    & 27.2 \\
\bottomrule
\end{tabular}%
}
\end{table*}

\paragraph{Additional TTS Results: Accuracy and Efficiency.}
We further evaluate GG against Best-of-N (BoN) and Chain-of-Thought (CoT) on AIME24, AIME25, MATH500, and AMC using \textit{DeepSeek-R1 1.5B and 7B} models (Table~\ref{tab:tts_accuracy_final}). To ensure fair and reliable comparison, particularly for AIME benchmarks known for high variance, all configurations were run four times with different seeds, reporting average and maximum accuracies across different runs. Experimental settings match those in Table~\ref{tab:model_performance_metrics_custom_order}.

Table~\ref{tab:tts_accuracy_final} demonstrates that GG consistently matches or surpasses BoN across most benchmarks and model sizes, particularly in mean accuracy, while BoN occasionally equals GG only in maximum scores. For example, on AIME24 with the 7B model, BoN shows some advantage in specific scenarios, but otherwise, GG maintains a strong comparative performance. Crucially, GG achieves these competitive or superior results against BoN while using approximately \textbf{50\% less KV cache memory}, a major efficiency gain. 

\section{Ablation Studies}
\label{sec:results_ablate}
We conduct several ablation studies to verify the efficacy of the different components that constitute GG. Due to space constraints, we focus here on the importance of RL fine-tuning with confidence, providing additional hyperparameter ablations (novelty weight and search) in the appendix.

\begin{wraptable}{Rt}{0.42\textwidth}
    \centering
    \caption{Effectiveness of RL fine-tuning onr AIME24: Performance scores with various reward strategies versus a no-RL baseline. Best result in bold.}
    \label{tab:ablation_reward_wraptab} 
    \begin{tabular}{lr} 
        \toprule
        Fine-tuning Setting      & Score $\uparrow$ \\
        \midrule
        No RL Fine-tuning        & 54.5\%  \\
        Correctness Reward Only  & 54.9\%  \\
        Confidence (No Penalty)  & 54.0\%  \\
        Confidence Reward (Ours) & \textbf{58.9}\%  \\
        \bottomrule
    \end{tabular}
\end{wraptable}
\paragraph{Impact of Confidence-Based RL Fine-tuning.}
To isolate the effectiveness of our proposed confidence-based reward mechanism within the RL fine-tuning phase (Equation~\ref{eq:conf_reward}), we conduct an ablation study comparing four settings: (1) without RL fine-tuning, (2) with RL fine-tuning using an accuracy reward, (3) with RL fine-tuning using a confidence reward (ours), and (4) RL fine-tuning using a confidence reward without penalty for incorrect answers.
Table~\ref{tab:ablation_reward_wraptab} summarizes our results which clearly demonstrate the importance of our confidence-based RL fine-tuning. 
Specifically, the full Confidence Reward method (58.9) substantially outperforms both the no RL fine-tuning baseline (54.5) and the fine-tuning strategy using only a Correctness Reward (54.9). Furthermore, removing the negative penalty for incorrect, highly confident answers results in the lowest score (54.0). This underscores the critical role of the penalty term for effective confidence calibration and confirms that the improvements stem directly from the specific reward design aimed at enhancing confidence calibration.

\section{Conclusion}
We introduce Guided by Gut (GG), a Test-Time Scaling framework enabling smaller Large Language Models (e.g., 1.5B parameters) to surpass significantly larger models (e.g., 32B) in performance while offering faster inference and substantially reduced GPU memory and KV cache usage. GG leverages intrinsic model signals—confidence and novelty—extracted directly from LLM outputs, further calibrated through reinforcement learning, providing a lightweight and effective alternative to costly external verifier-based methods such as Process Reward Models (PRMs). Compared to Best-of-N (BoN) strategies, GG achieves comparable or superior results with approximately 50\% less KV cache memory and competitive inference speeds, while delivering performance on par with PRM-guided approaches.

\paragraph{Limitations.}
\label{sec:Limitations}
The intrinsic signals employed in GG, such as confidence scores, do not inherently verify the actual correctness of each reasoning step. There exist cases where the model assigns high confidence to incorrect reasoning steps, hence the importance of our RL fine-tuning phase. Overall, 
the goal of our method is to provide a simple, efficient, and without relying on external verifier signals to guide the search process, rather than guaranteeing the absolute correctness of each step that may occur when using a strong PRM. We analyze and discuss representative failure cases in the Appendix to provide deeper insight into these limitations.

\paragraph{Broader Impacts.} 
\label{sec:BroaderImpacts}
This work proposes an efficient test-time reasoning framework for language models. Our contributions reduce the computational requirement needed to use reasoning models. Experimental results demonstrate that we can combine GG with a locally smaller LLM and achieve comparable performance to open-source models that require GPU rack servers or closed-source models behind an API. Therefore, one potential future impact of our work we hope to see is increased usage of locally-deployed reasoning LLMs.

\bibliographystyle{abbrvnat}
{
\small 
\bibliography{refs}
} 

\newpage
\appendix

\section{Supplementary Materials}
\label{sec:supp}
We provide additional details and analyses to complement the main paper. Section~\ref{sec:app_ablate} 
includes further ablation studies to dissect the contributions of individual components of our Guided by Gut (GG) framework. Section~\ref{sec:app_implementation} provides a detailed walkthrough of our Self-Guided Search algorithm. Finally, Section~\ref{sec:example} provides 
an illustrative example showcasing the step-by-step reasoning process of GG.

\subsection{Ablation Studies}
\label{sec:app_ablate}

In addition to the confidence-based RL fine-tuning ablation presented in the main paper (Section~\ref{sec:results_ablate}), we 
ablate the 
key components of GG. We specifically analyze the impact of the novelty weight ($\lambda_N$) in our reward formulation and the beam width ($M$) within the DVTS search algorithm.

\paragraph{Novelty: Method Selection and Weight ($\lambda_N$) Impact.}
For the novelty component $N(s^t)$, 
we consider two methods: The first is cosine similarity using embeddings from a \textbf{sentence transformer}. Sentence transformers are models that generate dense vector representations (embeddings) capturing sentence semantics; we specifically employed \texttt{all-MiniLM-L6-v2}~\cite{reimers2019sentence}. 
The second method is to count new words. 
As shown in Table~\ref{tab:ablation_novelty_method_resizebox}, counting new words performs 
comparably to cosine similarity but is simpler and computationally lighter. Thus, we selected word counting for $N(s^t)$.

Using word counting for novelty, we then investigate the impact of the corresponding weight, $\lambda_N$. Table~\ref{tab:ablation_novelty_resizebox} 
summarizes this ablation. Setting $\lambda_N = 0$ (no novelty signal) yields a score of 57.5. A balanced $\lambda_N = 0.5$ achieves the best score of 58.9. Conversely, setting $\lambda_N = 1$ (relying predominantly on novelty) significantly degrades performance to 51.9. This confirms that balancing novelty (for exploration) with confidence is crucial, with confidence being the primary guiding signal.

\paragraph{Impact of Beam Width ($M$).}
We examine the effect of the beam width ($M$) parameter within our DVTS search strategy, keeping the total path budget fixed at $N=32$. In DVTS, increasing $M$ reduces the number of independent subtrees ($N/M$) explored. We compare performance for $M=2$ (16 subtrees), $M=4$ (8 subtrees), and $M=8$ (4 subtrees), with results in Table~\ref{tab:ablation_beam_resizebox}. 
Increasing $M$ while keeping $N$ constant entails 
significantly worse performance: $M=2$ achieved 58.9, while $M=4$ dropped to 47.0, and $M=8$ further decreased to 44.0. This confirms that increasing $M$ under a fixed budget $N$ limits exploration diversity crucial for DVTS, making a smaller beam width ($M=2$) more effective for the $N=32$ budget.

\begin{table*}[t]
\centering
\begin{minipage}[t]{0.32\textwidth} 
    \centering
    \caption{Novelty method ablation on AIME24. Best Score bold.}
    \label{tab:ablation_novelty_method_resizebox}
    \resizebox{\linewidth}{!}{%
    \begin{tabular}{lr}
        \toprule
        Novelty Method & Score $\uparrow$ \\
        \midrule
        \textbf{New Token Counting} & \textbf{58.9} \\
        Cosine Similarity & 58.4 \\ 
        \bottomrule
    \end{tabular}
    }
\end{minipage}%
\hfill 
\begin{minipage}[t]{0.32\textwidth} 
    \centering
    \caption{Novelty weight ($\lambda_N$) ablation on AIME24 (New Token Counting). Best score bold.}
    \label{tab:ablation_novelty_resizebox}
    \resizebox{\linewidth}{!}{%
    \begin{tabular}{lr}
        \toprule
        Novelty Weight ($\lambda_N$) & Score $\uparrow$ \\
        \midrule
        0.0 & 57.5 \\
        0.5 & \textbf{58.9} \\
        1.0 & 51.9 \\
        \bottomrule
    \end{tabular}
    }
\end{minipage}%
\hfill 
\begin{minipage}[t]{0.32\textwidth} 
    \centering
    \caption{Beam width ($M$) ablation on AIME24. Total paths $N=32$. Best score bold.}
    \label{tab:ablation_beam_resizebox}
    \resizebox{\linewidth}{!}{%
    \begin{tabular}{lcr} 
        \toprule
        $M$ & Trees ($N/M$) & Score $\uparrow$ \\
        \midrule
        \textbf{2} & \textbf{16} & \textbf{58.9} \\
        4 & 8 & 47.0 \\
        8 & 4 & 44.0 \\
        \bottomrule
    \end{tabular}
    }
\end{minipage}
\end{table*}

\subsection{Implementation Details of Self-Guided Search}
\label{sec:app_implementation}

\begin{algorithm}[t]
\caption{\textbf{\textcolor{myblue}{Self-Guided Test-Time Scaling with Confidence-Calibrated DVTS}}}
\label{alg:self-guided-tts}

\KwIn{\textcolor{black}{Prompt $Q$, Model $\pi_{\theta}$, Beam width $M$, Total paths $N$, Max depth $T$, Token limit $\tau$}}
\KwOut{\textcolor{black}{Final Answer $A^*$}}
Initialize empty answer set $\mathcal{A}=\emptyset$\;
Initialize $\frac{N}{M}$ diverse subtrees from $Q$\;

\ForEach{subtree $j = 1$ to $N/M$ \textcolor{myorange}{\Comment*[r]{Traverse each subtree}}}{
Initialize path $R^{(j)} \leftarrow [\,]$\;

    \For{$t = 1$ \KwTo $T$ \textcolor{myorange}{\Comment*[r]{Roll out reasoning steps}}}{
        Generate $M$ candidate steps $\{s^t_i\}_{i=1}^{M}$ using model $\pi_{\theta}(R^{(j)})$\;

        \ForEach{candidate $s^t_i$ \textcolor{myorange}{\Comment*[r]{Score each candidate}}}{
          Compute $r_i = \lambda_C C(s^t) + \lambda_N N(s^t)$\; 
        }

        Select top-1 step $s^t_* \leftarrow \arg\max_i r_i$\;
        Append $s^t_*$ to $R^{(j)}$\;

        \If{$s^t_*$ contains final answer token (e.g., ``\texttt{boxed\{\}}'') \textcolor{mygreen}{\Comment*[r]{Answer is complete}}}{
            Extract $A^{(j)}$ and add to set $\mathcal{A}$\;
            \textbf{\textcolor{myred}{break and prune subtree $j$}}\;
        }

        \If{\texttt{TokenCount}($R^{(j)}$) $> \tau$ \textbf{or} $t = T{-}1$ \textcolor{mygreen}{\Comment*[r]{Force answer near limit}}}{
            Append ``\texttt{Final Answer}'' to $R^{(j)}$\;
        }
    }
}

Select final answer $A^* \leftarrow$ confidence-weighted vote over $\mathcal{A}$\;
\Return{$A^*$}
\end{algorithm}

To clarify how the search operates, we walk through Algorithm~\ref{alg:self-guided-tts} step by step. Our approach builds on Diverse Verifier Tree Search, a variant of beam search that partitions the total number of candidate paths $N$ into $\frac{N}{M}$ diverse subtrees. Each subtree is then greedily expanded based on confidence-calibrated intrinsic rewards.

The algorithm begins with a prompt $Q$, a language model $\pi_{\theta}$, and user-defined hyperparameters such as the total number of paths $N$. In line 2, the LLM is queried with $Q$ to generate initial reasoning branches, forming the roots of several subtrees. For each subtree, we define a maximum search depth $T$ and consider $M$ candidate next steps at each level. These candidates are scored using the reward in Eq.~\ref{eq:gg_reward} (line 11), and the highest-scoring step is appended to the current reasoning chain (lines 12–13).

The procedure continues recursively until one of the termination criteria (lines 14–20) is met: (1) the reasoning chain exceeds depth $T$, (2) the total token budget $\tau$ is surpassed, (3) the model exhibits signs of degeneration such as repetitive output, or (4) a complete reasoning chain $R$ with a conclusive final answer $A$ is generated. To mitigate premature truncation near token or depth limits, we incorporate model-specific prompts that encourage finalization. For instance, appending "Final Answer" is particularly effective with DeepSeek models for reliably triggering a conclusive output. Finally, Algorithm~\ref{alg:self-guided-tts} aggregates the candidate completions and selects the final answer $A^*$ using a confidence-weighted voting scheme defined in Eq.~\ref{eq:seq_confidence}.

\definecolor{darkgreen}{RGB}{0,100,0}

\subsection{Search and Self-Guidance Example} 
\label{sec:example}
We analyze a particular reasoning trace, distinct from the illustrative example shown in Fig.~\ref{fig:reasoning_strategies}, to demonstrate a scenario where an error initially occurred and to highlight the step-by-step operation of the Guided by Gut (GG) framework.

Even though intrinsic confidence often serves as a reliable guiding signal, the model can still make mistakes, as it lacks a mechanism to definitively verify correctness. In the provided example trace, initially, the high-confidence Branch 2 mistakenly computes the sum as 459 pounds, despite a confidence score of 0.89 at Step 2 and 0.79 at Step 5. However, the intrinsic confidence eventually leads to a self-correction: at Step 7, Branch 2 corrects its previous error, accurately computing the sum as 449 pounds with a confidence of 0.82. From this point forward, the correct answer is consistently maintained.

This behavior highlights the GG framework’s key strength: it effectively leverages intrinsic signals from the model to guide reasoning decisions with negligible computational overhead, achieving performance comparable to other Test-Time Scaling (TTS) methods like Process Reward Models (PRMs), but crucially without the heavy computational demands typically associated with them.

\medskip

\begin{tcolorbox}[colback=white, colframe=black, title={Question and the Reasoning Trace}, breakable, boxrule=1pt]
\noindent\textbf{Question:} In the quiet town of \emph{Balancing Scales}, four friends like to cloak their weights in puzzles.
\begin{itemize}
  \item Ivy confides, “If you double my weight and then add a baker’s dozen, you’ll land exactly on Jasper’s number.”
  \item Jasper counters, “Kai is as heavy as the arithmetic mean of Ivy and me.”
  \item Kai chuckles, “That may be so, yet Luna still comes in eight pounds lighter than I do.”
\end{itemize}
Everyone already knows that Ivy tips the scale at 125 pounds. \textbf{What is the combined weight of Jasper and Luna, expressed in pounds?}

\begin{center}\textbf{Step 2}\end{center}
\medskip

\begin{itemize}
  \item \textcolor{darkgreen}{\textbf{Branch 1, Confidence Score: 0.88:}}
    \begin{itemize}
      \item 2. \textbf{Jasper and Kai's Weight Relationship}:
      \item Jasper says, "Kai weighs the average of Ivy and me."
      \item Let me think about this. If I take the average of two numbers, that's their sum divided by two. So, mathematically:
            \( \text{Kai's weight} = \frac{\text{Ivy's weight} + \text{Jasper's weight}}{2} \)
      \item Plugging in the known values:
            \( \text{Kai's weight} = \frac{125 + 263}{2} \)
      \item Let's compute the numerator first:
            \( 125 + 263 = 388 \)
      \item Then divide by 2:
            \( 388 / 2 = 194 \)
      \item So, Kai weighs 194 pounds. Got that down.
    \end{itemize}

  \item \textcolor{red}{\textbf{Branch 2, Confidence Score: 0.89:}}
    \begin{itemize}
      \item 2. \textbf{Jasper and Kai's Weight Relationship}:
      \item Jasper says, "Kai weighs the average of Ivy and me."
      \item This means that Kai's weight K is the average of Ivy's and Jasper's weights.
      \item So, mathematically:
            \( K = \frac{\text{Ivy's weight} + \text{Jasper's weight}}{2} \)
      \item Plugging in the known values:
            \( K = \frac{125 + 263}{2} \)
      \item Let me compute the numerator: \(125 + 263 = 388\)
      \item Then divide by 2: \(388 / 2 = 194\)
      \item So, Kai is 194 pounds.
      \item 3. \textbf{Kai and Luna's Relationship}:
      \item Kai laughs, “That’s true, but Luna still comes in eight pounds lighter than I am.”
      \item This tells us that Luna's weight L is equal to Kai's weight minus 8 pounds.
            \( L = K - 8 \)
      \item Since we found K as 194 pounds, this gives:
            \( L = 194 - 8 = 186 \) pounds.
      \item So, Luna is 186 pounds.
      \item 4. \textbf{We need to find the combined weight of Jasper and Luna}:
      \item So, that's \( J + L \).
      \item We already know J is 263 and L is 186, so adding those together:
      \item \(263 + 186 = \dots\) Let me compute that:
      \item 200 + 100 = 300
      \item 60 + 80 = 140
      \item 3 + 6 = 9
      \item Wait, that might not be the right way to break it down. Alternatively, just adding them directly:
      \item \(263 + 186\):
      \begin{itemize}
        \item Start with the units place: \(3 + 6 = 9\)
        \item Tens place: \(6 + 8 = 14\), carryover 1
        \item Hundreds place: \(2 + 1 = 3\), plus the carried-over 1 makes 4
      \end{itemize}
      \item \textcolor{red}{So altogether, 459 pounds.}
    \end{itemize}
\end{itemize}

\begin{center}\textbf{Step 3}\end{center}
\medskip

\begin{itemize}
  \item \textcolor{darkgreen}{\textbf{Branch 1, Confidence Score: 0.5:}} Let me double-check just in case I made any arithmetic mistakes.
  \item \textcolor{darkgreen}{\textbf{Branch 2, Confidence Score: 0.66:}} Wait, let me verify each step again to make sure I didn't make an arithmetic error.
\end{itemize}

\begin{center}\textbf{Step 5}\end{center}
\medskip

\begin{itemize}
  \item \textcolor{darkgreen}{\textbf{Branch 1, Confidence Score: 0.73:}}
    \(263 + 186\):
    Starting from the right:
    \begin{itemize}
      \item \(3 + 6 = 9\)
      \item \(6 + 8 = 14\), carryover 1
      \item \(2 + 1 + 1 \text{(carried)} = 4\)
    \end{itemize}
    So, yes, 449. Previously, I thought \(388/2\) was 194, then \(194-8=186\), and then adding \(263+186\).
  \item \textcolor{red}{\textbf{Branch 2, Confidence Score: 0.79:}}
    Compute \(263 + 186\):
    \begin{itemize}
      \item Units place: \(3 + 6 = 9\)
      \item Tens place: \(6 + 8 = 14\), write down 4, carryover 1
      \item Hundreds place: \(2 + 1 = 3\), plus the carried-over 1 makes 4
    \end{itemize}
    So total is 459. So why did I think it was 449 earlier? Maybe a mistake in initial breakdown.
\end{itemize}

\begin{center}\textbf{Step 6}\end{center}
\medskip

\begin{itemize}
  \item \textcolor{darkgreen}{\textbf{Branch 1, Confidence Score: 0.6:}} Wait, no, my breakdown seems correct: \(400 + 40 + 9 = 449\). Oh! Wait, wait—no. \(263 +186\): Maybe an error in adding. Let's try adding \(263 + 186\) step by step.
  \item \textcolor{red}{\textbf{Branch 1, Confidence Score: 0.52:}} Wait, \(263 +186\): I think I miscalculated the tens and hundreds in my head. Maybe it's 459, correct.
\end{itemize}

\begin{center}\textbf{Step 7}\end{center}
\medskip

\begin{itemize}
  \item \textcolor{red}{\textbf{Branch 1, Confidence Score: 0.68:}}
    Starting from the right:
    \begin{itemize}
      \item 3 (units place) + 6 (units) = 9
      \item 6 (tens place) + 8 (tens) = 14, carryover 1
      \item 2 (hundreds place) + 1 (hundreds) = 3, plus the carried over 1 = 4
    \end{itemize}
    So, it's 459. So correct.
  \item \textcolor{darkgreen}{\textbf{Branch 2, Confidence Score: 0.82:}}
    \(263 + 186\). Break it down:
    \begin{itemize}
      \item Start with 263. Let me add 100 to it: \(263 + 100 = 363\)
      \item Then add 80: \(363 + 80 = 443\)
      \item Then add 6: \(443 + 6 = 449\)
    \end{itemize}
\end{itemize}

\begin{center}\textbf{Step 8}\end{center}
\medskip

\begin{itemize}
  \item \textcolor{darkgreen}{\textbf{Branch 1, Confidence Score: 0.81:}}
    Wait, so if I add \(263 + 186\):
    \begin{itemize}
      \item \(200 + 100 = 300\)
      \item \(60 + 80 = 140\)
      \item \(3 + 6 = 9\)
    \end{itemize}
    Total: \(300 + 140 + 9 = 449\).
  \item \textcolor{red}{\textbf{Branch 2, Confidence Score: 0.45:}}
    Wait earlier, when I thought 449, I was off by one. No, wait—no, actually \(263 + 186 = 449\) is incorrect. So where is the confusion? Is it an arithmetic error?
\end{itemize}

\begin{center}\textbf{Step 18}\end{center}
\medskip

\begin{itemize}
  \item \textcolor{darkgreen}{\textbf{Branch 1, Confidence Score: 0.73:}}
    Wait, so earlier, when I thought it was 459, that was a mistake.
  \item \textcolor{darkgreen}{\textbf{Branch 2, Confidence Score: 0.67:}}
    Wait, so that's the correct combined weight. So earlier when I thought 459, that was a miscalculation. So now, the answer is 449 pounds.
\end{itemize}
\end{tcolorbox}

\end{document}